\documentclass[runningheads]{llncs}

\usepackage[T1]{fontenc}
\usepackage{graphicx}

\usepackage{wrapfig}
\usepackage{graphicx}
\usepackage{amsmath}
\usepackage{nameref}
\usepackage{caption}
\usepackage{subcaption}
\usepackage{color}
\usepackage{amsfonts}
\usepackage{multirow}
\usepackage{soul}
\usepackage{algorithm}
\usepackage{algpseudocode}
\usepackage{xspace}
\usepackage[dvipsnames]{xcolor}
\usepackage{colortbl}
\usepackage{hyperref}
\usepackage[nameinlink]{cleveref}
\usepackage[misc,geometry]{ifsym}

\newcommand{\algo}{\texttt{DALLMi}\xspace}

\begin{document}

\title{\algo: Domain Adaption for LLM-based Multi-label Classifier}

\author{
Miruna Bețianu\inst{1}\orcidID{0009-0009-1010-0148}
\and
Abele M\u{a}lan\Letter\inst{1,2}\orcidID{0009-0002-4493-7439}
\and
Marco Aldinucci\inst{3}\orcidID{0000-0001-8788-0829}
\and
Robert Birke\inst{3}\orcidID{0000-0003-1144-3707}
\and
Lydia Chen\inst{1,2}\orcidID{0000-0002-4228-6735}
}

\authorrunning{M. Bețianu et al.}

\institute{
Delft University of Technology, Delft, Netherlands. \\
Email: \email{m.betianu@student.tudelft.nl} \\
\and
University of Neuch\^atel, Neuch\^atel, Switzerland. \\
Email: \email{\{abele.malan, yiyu.chen\}@unine.ch} \\
\and
University of Turin, Torino, Italy. \\
Email: \email{\{marco.aldinucci, robert.birke\}@unito.it}
}

\maketitle

\begin{abstract}

Large language models (LLMs) increasingly serve as the backbone for classifying text associated with distinct domains and simultaneously several labels (classes). When encountering domain shifts, e.g., classifier of movie reviews from IMDb to Rotten Tomatoes, adapting such an LLM-based multi-label classifier is challenging due to incomplete label sets at the target domain and daunting training overhead. The existing domain adaptation methods address either image multi-label classifiers or text binary classifiers. In this paper, we design \algo, \textbf{D}omain \textbf{A}daptation \textbf{L}arge \textbf{L}anguage \textbf{M}odel \textbf{i}nterpolator, a first-of-its-kind semi-supervised domain adaptation method for text data models based on LLMs, specifically BERT. The core of \algo is the novel variation loss and MixUp regularization, which jointly leverage the limited positively labeled and large quantity of unlabeled text and, importantly, their interpolation from the BERT word embeddings. \algo also introduces a label-balanced sampling strategy to overcome the imbalance between labeled and unlabeled data. We evaluate \algo against the partial-supervised and unsupervised approach on three datasets under different scenarios of label availability for the target domain. Our results show that \algo achieves higher mAP than unsupervised and partially-supervised approaches by 19.9\% and 52.2\%, respectively.


\end{abstract}

\section{Introduction}

Text classification is a fundamental task in Natural Language Processing with a diverse range of applications, including sentiment analysis~\cite{nasukawa2003sentiment}, spam detection~\cite{crawford2015survey}, and document categorization~\cite{fatima2017text}. Large Language Models (LLMs) are the state-of-the-art backbone for such classification tasks~\cite{sun2023text}. By pre-training on large and heterogeneous corpora, they achieve good baseline performance across various domains, even in cases where multiple labels\footnote{each label represents a possible class} may be present simultaneously. When encountering a new domain with a sufficient amount of labeled data, fine-tuning LLMs can further increase the accuracy of classification tasks.

\begin{figure}[tb]
    \centering
    \begin{minipage}{0.38\textwidth}
        \centering
        \includegraphics[width=\textwidth]{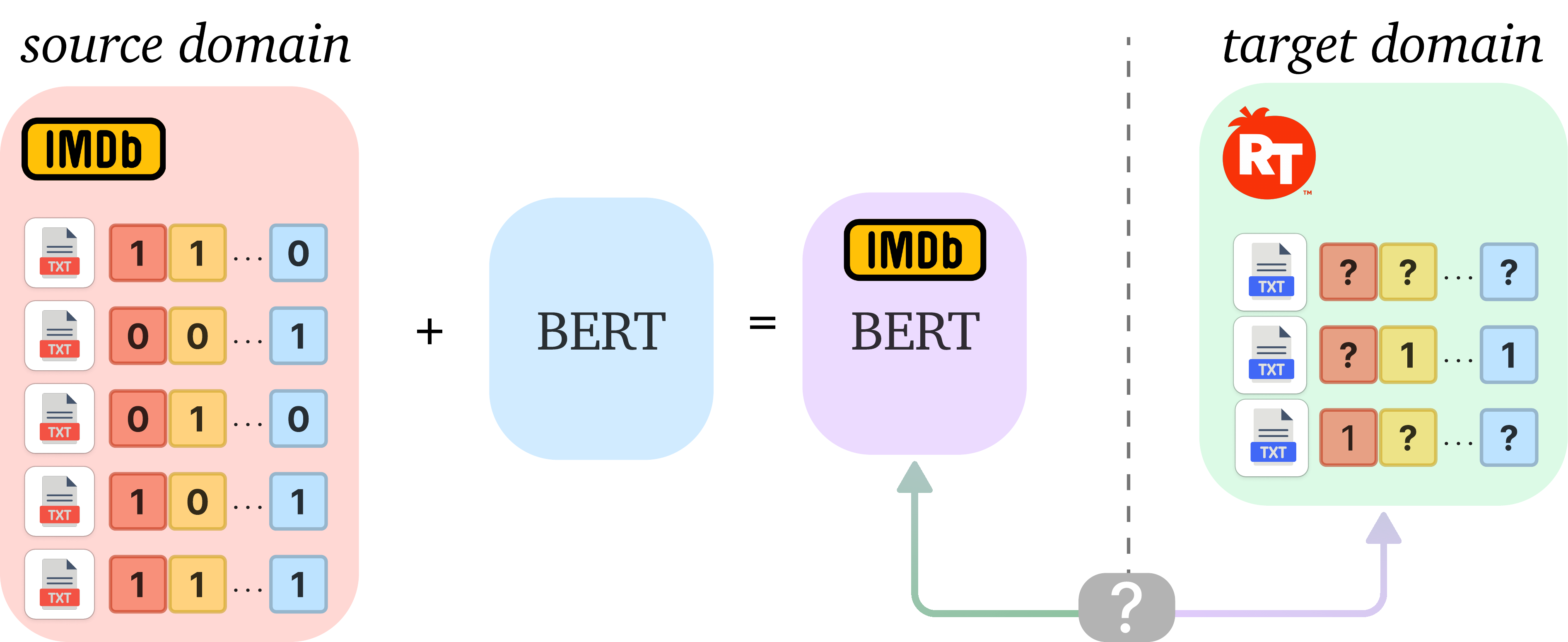}
        \caption{Adapting \textit{BERT} from IMDb to Rotten Tomatoes.}
        \label{fig:problem-statement}
    \end{minipage}
    \hfill
    \begin{minipage}{0.58\textwidth}
        \centering
        \includegraphics[width=\textwidth]{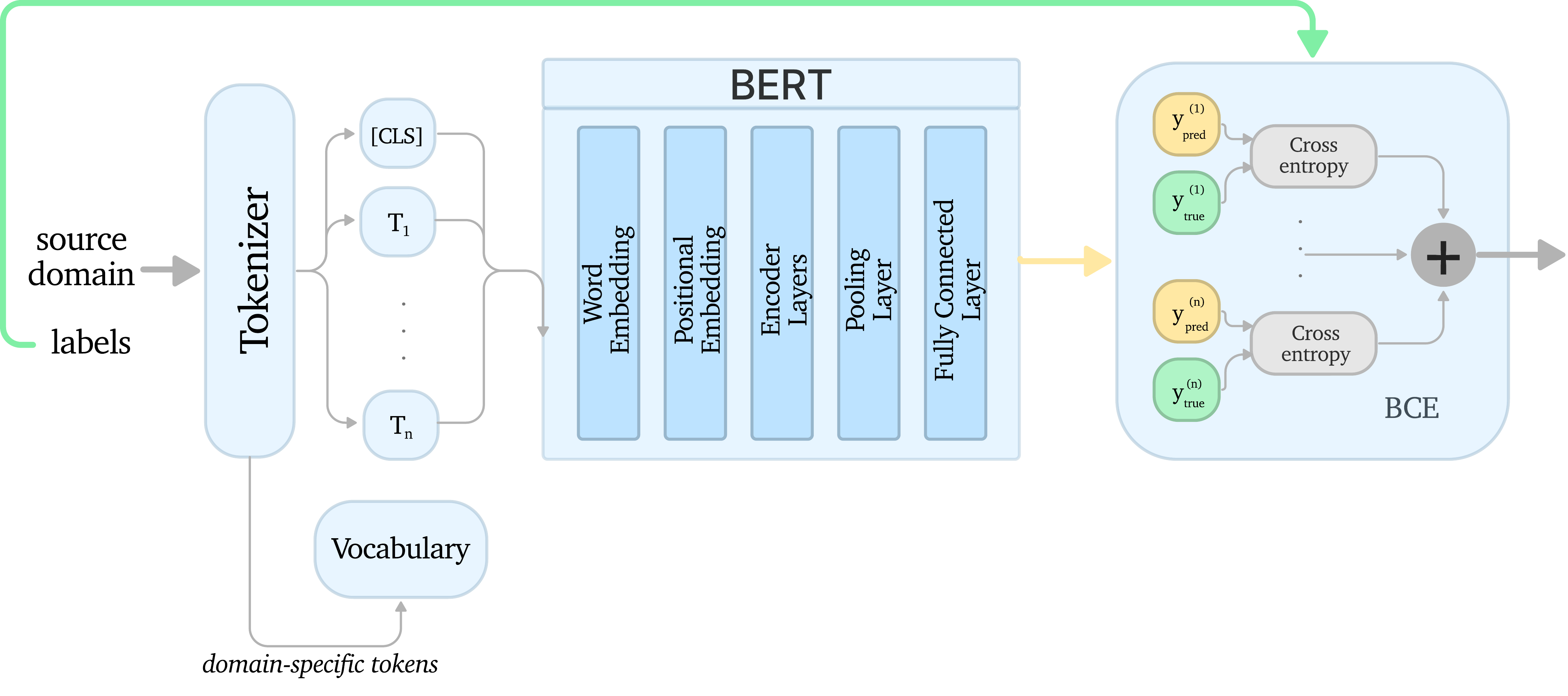}
        \caption{\textit{BERT} multi-label classification flow.}
        \label{fig:bert-mlc}
    \end{minipage}
\end{figure}

In real life scenarios, there is only limited labeled data in many target domains, making regular fine-tuning ineffective and requiring a domain shift from source domain. \autoref{fig:problem-statement} outlines an example of such a challenge in classifying text into multiple labels/classes. Assume we have an LLM trained to predict the genres of a movie from an IMDb dataset, as the source domain. Each movie is associated with multiple genres/labels. We then have to perform the same task on a new dataset from Rotten Tomatoes, as the target domain, which contains incomplete labels. The annotation of the target domain is a time-consuming and resource-intensive task, nonetheless unfeasible. The research challenge is how to utilize the knowledge of the source domain and the limited labels for the target domain to enhance the performance of LLM-based multi-label classifiers.

Existing text classification adaptation research~\cite{singhal2023domain} focuses on the multi-class case, where each sample corresponds to only one label from the available set, like in the ubiquitous sentiment analysis task or the even simpler binary one. The multi-label scenario, where samples correspond to any number of labels, poses an increased challenge, further increased by possibly underrepresented labels in the available data. Most prior multi-label domain adaptation efforts lie within the modality of image data~\cite{9434003,singh2023discriminatorfree}. Moreover, only few unsupervised approaches exist to finetune LLMs~\cite{ryu2020knowledge,rietzler2019adapt}, which are usually limited to single label classification.

This paper introduces \algo, a novel semi-supervised technique for LLM adaptation to different textual domains. We combine supervised fine-tuning on a source text domain with semi-supervised fine-tuning on a partially labeled text target domain. Specifically, we focus on scenarios where there are limited positive samples for each label, and the remaining samples are unlabeled, as shown in \autoref{fig:problem-statement}. \algo introduces a variational loss that leverages labeled and unlabeled information to maximize the knowledge extracted from all samples. To further boost its classification ability, \algo augments the target dataset with synthetic samples generated by mixing labeled and unlabeled ones. We extensively evaluate \algo under various settings for multiple datasets, including ablations on relevant features. To recap, we make the following contributions:

\begin{itemize}
\item We design \algo, the first semi-supervised LLM domain adaptation framework for multi-label text classification.
\item We design a novel variational loss for the model to learn from both labeled and unlabeled parts of target instances.
\item We interpolate representations of labeled and unlabeled samples into new synthetic ones to regularize our variational loss.
\item \algo outperforms partial fine-tunning and unsupervised approaches by 19.9\% and 52.2\%, respectively.
\item We make the code available at \url{https://github.com/mirunabetianu/DALLMi}.
\end{itemize}

\section{Language Model and Domain Adaptation}

\textbf{BERT for multi-label classification.}
\autoref{fig:bert-mlc} shows a high-level overview of the architecture of BERT's multi-label classifier, which assumes a full set of labels. The input sentence is initially passed through the tokenizer to obtain tokens, paddings, and truncation to prepare it for BERT. This process may also include the addition of new tokens to BERT's vocabulary. The encoded sentence proceeds through BERT's forward method, which incorporates embedding and encoding layers, to obtain the internal hidden representations. A pooling layer decreases the hidden representation's dimension, from which a classification head (usually a single fully connected layer) obtains the final logits. An activation function may be applied to determine predicted labels based on a threshold. The default training approach uses multiple binary cross entropy losses to treat the multi-label classification as several binary problems.

\textbf{Prior Art Domain Adaptation.}
Domain adaptation addresses the dataset shift~\cite{quinonero2008dataset} between the source and target domains. It can be split into four main categories: fully supervised, semi-supervised, weakly supervised, and fully unsupervised. Fully supervised methods rely heavily on many target labels, aiming to generalize the model to both domains~\cite{DBLP}. When the target domain labels are limited, the problem becomes semi-supervised. Common approaches combine source and domain data to improve model training~\cite{zhang2021semisupervised}. Weakly supervised methods aim to adapt the model by considering the uncertainty and limitations of the target domain~\cite{Deng2021AWS}. Unsupervised domain adaptation aims to adapt the model without any information about the target labels, approaches including batch normalization tuning~\cite{wang2021tent}, feature alignment \cite{eastwood2022sourcefree}, or whole network training~\cite{learningtoadapt2020}. Most methods operate within image domains. Nonetheless, some techniques exist for LLMs, with fine-tuning being the most common adaptation method~\cite{grangier-iter-2022-trade,guo-etal-2021-bertweetfr,Buonocore_2023}. Other methods expand the LLM’s vocabulary, enhancing the corpus with specific domain tokens~\cite{sachidananda-etal-2021-efficient}, or its architecture, adding new adapter layers~\cite{chronopoulou2022efficient}. In terms of classification adaptation, the majority of LLM-based approaches are designed for sentiment analysis, using adversarial adaptation with knowledge distillation~\cite{rietzler2019adapt} or supervised fine-tuning~\cite{ryu2020knowledge}. Our method, however, targets general classification workloads and employs semi-supervised fine-tuning.

\section{\algo}

Adapting a fine-tuned LLM to partially labeled target data is a challenging task, primarily due to the need for a substantial amount of labels to ensure that the model gives reliable predictions. To address the challenge posed by the scarcity of the target domain, we introduce a new semi-supervised fine-tuning method based on Positive Unlabeled (PU) learning and data augmentation. Our method takes inspiration from VPU~\cite{chen2020variational}, estimating separate losses for positive and unlabeled samples and combining positive-unlabeled pairs to generate artificial samples to compensate for the limited number of labels.

Formally, we aim to learn a multi-label classifier, $\Phi: X^t \rightarrow Y$, where $X^t \in \mathbb{R}^d$ is the input data from the target domain, and $Y=[y_1 \dots y_L], \forall y_l\in [1,0]$ is the value vector for each label from 1 to $L$. Given some label $l$, $y_l=1$ denotes the presence of a positive example, while $y_l=0$ denotes the absence of an annotation. For each possible label $l\in \{0 \dots L\}$, we separate $X^t$ into two sets: the unlabeled set $\mathcal{U}_N^l$ with samples $s_u$ and the positive set $\mathcal{P}_N^l$ with samples $s_p$.

\subsubsection{Primer on VPU.}

Positive unlabeled learning~\cite{Bekker_2020} focuses on the scenarios where labeled samples are always a subset of the positive ones, and unlabeled ones may be positive or negative. The goal of the trained model is to distinguish between positive and negative instances, as in the classic fully labeled case~\cite{Bekker_2020}. Rooted in the variational principle, VPU~\cite{chen2020variational} uses a novel variation loss function to approximate the ideal \textbf{binary} classifier without the class prior, given unlabeled and positive samples. The variational loss function aims to quantify the empirical difference between the predicted outcomes for unlabeled and positive samples. In addressing the lack of positive samples, the VPU algorithm introduces a consistency regularisation term that incorporates the principles of MixUp, mitigating the problem of overfitting and increasing the robustness of the model. The MixUp term uses data augmentation techniques to generate synthetic samples by interpolating between positive and unlabeled instances. In addition, the algorithm quantifies the consistency between the model's predictions and the predictions made on these interpolated samples by the mean squared logarithmic error. We note that PU-MLC considers the problem of expanding VPU for multi-label classification~\cite{yuan2023positive}, but in the image analysis domain.

\subsubsection{Overview of \algo.}

To jointly leverage both positive labeled and unlabeled data, \algo fine-tunes the LLM-based multi-label classifier in a semi-supervised manner. We extend the design idea of VPU, combining the variation loss per label and MixUp regularization based on data interpolation, for a multi-label classifier. In contrast to image data or tabular data, such an interpolation is not straightforward for text data, which is represented as tokens in the LLM. We thus need to design a novel loss function and interpolation scheme to compute the MixUp, and then train the multi-label classifier and backbone BERT.

\begin{figure}[tb]
    \centering
    \includegraphics[trim={0 0.25cm 0 0},clip,width=\textwidth]{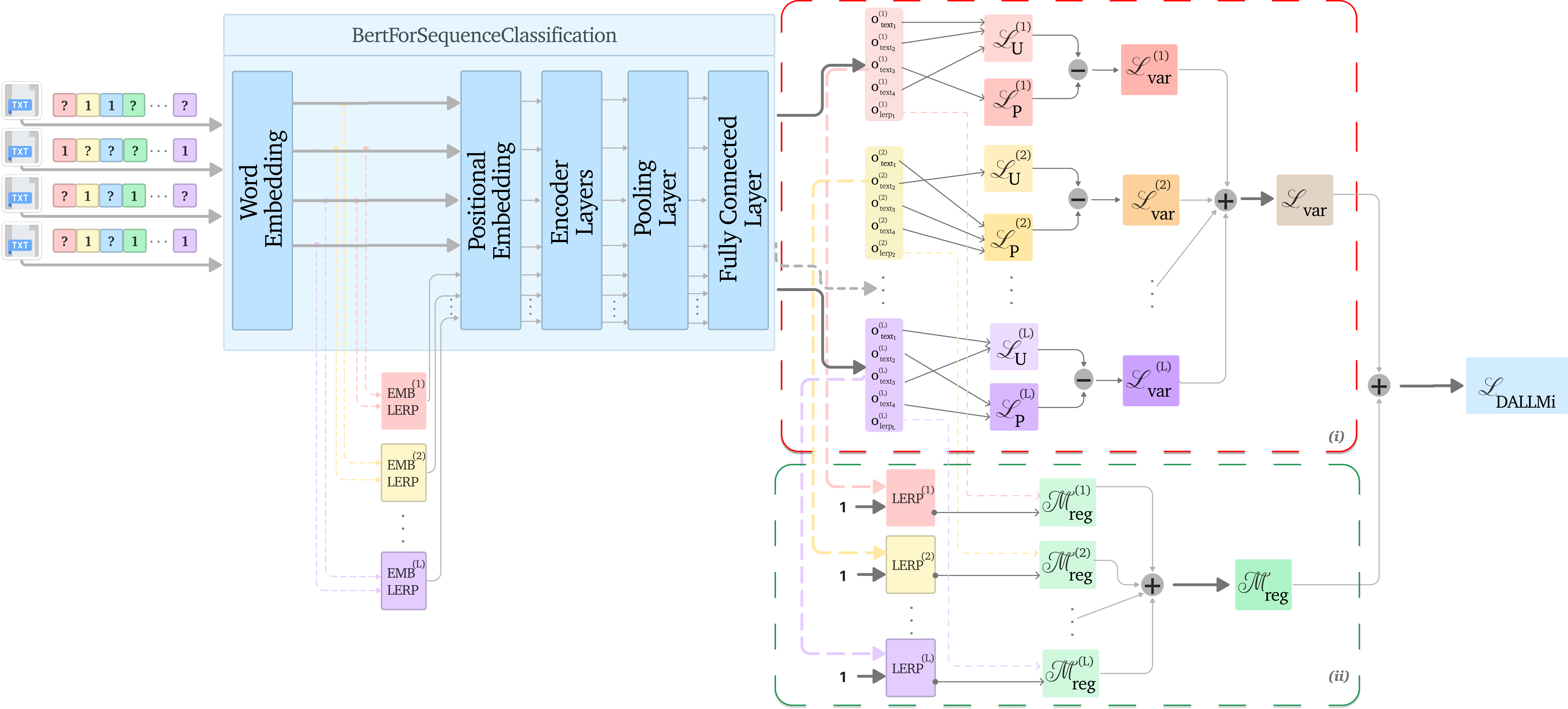}
    \caption{\algo flow: Unlabeled ($?$) and positive ($1$) samples from the target domain are fed through BERT, generating label-specific output logits. The logits are used to compute partial per-label variational losses for unlabeled and positive samples (dashed box (i)). The MixUp regularization combines per label linear interpolations (LERPs) applied to both inputs and outputs (dashed box (ii)).}
    \label{fig:overview}
\end{figure}

\algo consists of three key novelties, namely label-balanced sampling, variational loss and MixUp regularization. \autoref{fig:overview} shows the primary flow of \algo for computing the proposed variational loss and MixUp regularization for an example set of 4 samples. The target domain is pre-processed to ensure compatibility with the BERT classifier (blue shaded box). After the initial word embeddings layer, a distinctive input interpolation mechanism is introduced for each label. Here, unlabeled and positive embeddings are paired to generate new artificial embeddings via a linear interpolation block (LERP). After interpolation, the original and newly generated embeddings pass through the remaining layers of the BERT classifier. The final layer features a classification head represented as a single fully connected layer outputting $L$ logit scores. These logit scores are used to compute a specialised multi-label variational loss, $\mathcal{L}_{var}$, which sums up all per-label variation losses (dashed block (i)). The MixUp regularization $\mathcal{M}_{reg}$ sums up per label combinations of input interpolations and output interpolations, which are determined by linearly interpolating the logits derived from the input interpolation and the true output (dashed block (ii)).

\subsubsection{Label-balanced Sampling.}

In multi-label classification, where labels are representable as a sparse matrix, the challenge is creating label-balanced batches, particularly as the label count increases. Since, for any label there are often more unlabeled samples than positive ones, we need to overcome this disparity by ideally boosting the number of labeled samples from each label. Therefore, the first step in \algo involves an efficient and balanced sampler that ensures at least one positive sample from each label exists in every batch.

To achieve such balance, we introduce a cycle sampler that iterates through every label and retrieves one positive sample, repeating the process as many as needed until to reach the desired batch size. By ensuring the presence of at least one positive sample for each label, we obtain a fair and accurate estimation of the variational loss and allow calculating the linear interpolation for the MixUp.

\subsubsection{Variational Loss per Label.}

Having at least one positive sample from each label allows the loss function to make reasonable estimations across the sample set. In this context, we introduce the variational loss; see the red dashed box in \autoref{fig:overview}. For each label, the loss, grounded in the variational principle, uses positive and unlabeled samples to minimize the divergence between the positive distributions of the modeled and ideal classifiers as a proxy for the original task. The difference between the log of the expected sigmoids for unlabeled data and the expected value of the log sigmoids for labeled data serves as the approximation for the classifier's positive distribution. Interestingly, through empirical analysis, our findings diverge from this conventional approach. Specifically, we observe that replacing the log operations with the norm yields significant performance improvements. This norm-based approach is particularly advantageous when working within a specific domain. In addition, using the logarithm can exacerbate errors in certain situations, especially when dealing with labels with very low probabilities. This problem becomes more pronounced as the width of the datasets increases. In both cases, aggregating the values across all possible labels gives the final variational loss:
\begin{equation}
    \mathcal{L}^{l}_{var} (\Phi) = \frac{1}{|\mathcal{U}^{l}_{N}|} \sum _{s_u \in \mathcal{U}^{l}_{N}} \sigma (\Phi(s_u))
                 - \frac{1}{|\mathcal{P}^{l}_{N}|} \sum _{s_p \in \mathcal{P}^{l}_{N}} || \sigma (\Phi(s_p)) ||,
\label{eq:var-loss}
\end{equation}
where $\sigma (\cdot)$ is the sigmoid function, and $|| \cdot ||$ represents the norm of a vector.

Although the effectiveness of variational loss is proven~\cite{chen2020variational,yuan2023positive}, it has limitations, notably the need for a substantial number of positive samples in the dataset. To mitigate this limitation, we use data augmentation.

\subsubsection{MixUp Regularization.}

We compute the MixUp~\cite{zhang2018mixup,pmlr-v97-verma19a} term based on synthetically augmented data. The primary purpose of the MixUp is to generate additional samples by interpolating between positive and unlabeled samples, thus compensating for the lack of positive labels. The interpolations are weighted combinations using the Beta distribution to facilitate a smooth transition. In addition to effectively expand the dataset size, the MixUps quantify the consistency between the predicted and interpolated outputs.

LLM inputs consist of sequences of tokens that are not necessarily closer in value if they are closer in meaning. Therefore, simple token-level interpolation could result in problematic combinations that may hinder the model's learning process.
For the interpolation to be possible, we need to delve into the BERT model layers to identify which internal representations of the text data would be suitable~\cite{pmlr-v97-verma19a}. \autoref{fig:MixUp} illustrates three possible locations where to extract internal representations to perform MixUps and generate artificial samples. The first location is after the tokens pass the embedding layer of BERT. We extract the word embeddings for the chosen unlabeled and positive sentences and perform interpolation using the Beta distribution (see \autoref{eq:MixUp}). The resulting new embedding undergoes the subsequent layers of BERT to generate predictions for this new sample. Similarly, the other two locations extract, interpolate and feed back the embeddings at the deeper encoding and pooling layers of the model. Our empirical evaluation of the all interpolation locations favours the word embedding variant. We then compute the MixUp via a norm-based mean squared error to determine the consistency between the predicted outcome for unlabeled samples and of interpolated one:
\begin{align}
    \mathcal{M}^{l}_{reg} (\Phi) &= ( || \sigma (\Tilde{\Phi}) ||  - || \sigma (\Phi ( \Tilde{e})) || ) ^2,\label{eq:MixUp} \\
    \text{where } \mu &\stackrel{iid}{\sim} Beta (\alpha, \beta)\\
    \Tilde{e} &= \mu \cdot e_p + (1 - \mu) \cdot e_u \label{eq:embedding-lerp}\\
    \Tilde{\Phi} &= \mu \cdot 1 + (1 - \mu) \cdot \Phi(s_u)\label{eq:output-lerp}
\end{align}
where $e_p$ and $e_u$ represent the embedding corresponding to a positive $s_p$ and unlabeled $s_u$ sample, respectively; the hyperparameters $\alpha$ and $\beta$ control the shape of the Beta distribution.

\autoref{alg} summarises the overall computation of \algo loss. Following the steps displayed in \autoref{fig:overview}, a balanced batch is selected and used for the calculation of per-label variational loss. Additionally, the algorithm creates input and output interpolations to compute per-label MixUp values. The last step of the algorithm combines the variational loss and MixUp regularization terms weighted via the $\lambda$ hyperparameter. The gradients of the final loss are computed using backpropagation, which are then used to update the BERT classifier parameters and weights, gradually guiding the model towards better performance.

\begin{algorithm}[tb]
\caption{\algo{} -- Loss Calculation}
\label{alg}
\textbf{Input}: Label-balanced sampled data batch $B$, BERT model $\Phi$, hyperparameters $\alpha$/$\beta$/$\lambda$\\
\textbf{Output}: Loss value for samples in $B$
\begin{algorithmic}[1]
\For{each label $l$}
        \State Select unlabeled samples $U^l_N$ and positive samples $P^l_N$ from batch $B$
        \State Calculate variational loss by \autoref{eq:var-loss}
    \State Select one pair of unlabeled and positive inputs, $s_u$ and $s_p$
    \State Sample $\mu \sim Beta(\alpha, \beta)$
    \State $e_u$, $e_p$ = $\Phi$.embeddings($s_u$), $\Phi$.embeddings($s_p$)
    \State Compute interpolation by \autoref{eq:embedding-lerp}
    \State Compute artificial model prediction by \autoref{eq:output-lerp}
    \State Compute MixUp term by \autoref{eq:MixUp}
\EndFor
\State Compute the overall loss $\mathcal{L}_\text{final}(\Phi) = \sum _{l=1}^{L} \mathcal{L} _{var}^{l}(\Phi) + \lambda \mathcal{M} _{reg}^{l}(\Phi)$ \label{eq:final-loss}
\State Backpropagate $\mathcal{L}_\text{final}(\Phi)$
\end{algorithmic}
\end{algorithm}

\section{Experiments}

We assess the effectiveness of \algo against different LLM fine-tuning methods on three multi-label datasets: \textit{PubMed}, \textit{ArXiv}, and \textit{Movies}.

\begin{figure}[tb]
    \centering
    \begin{minipage}{0.44\textwidth}
        \centering
        \includegraphics[width=\textwidth]{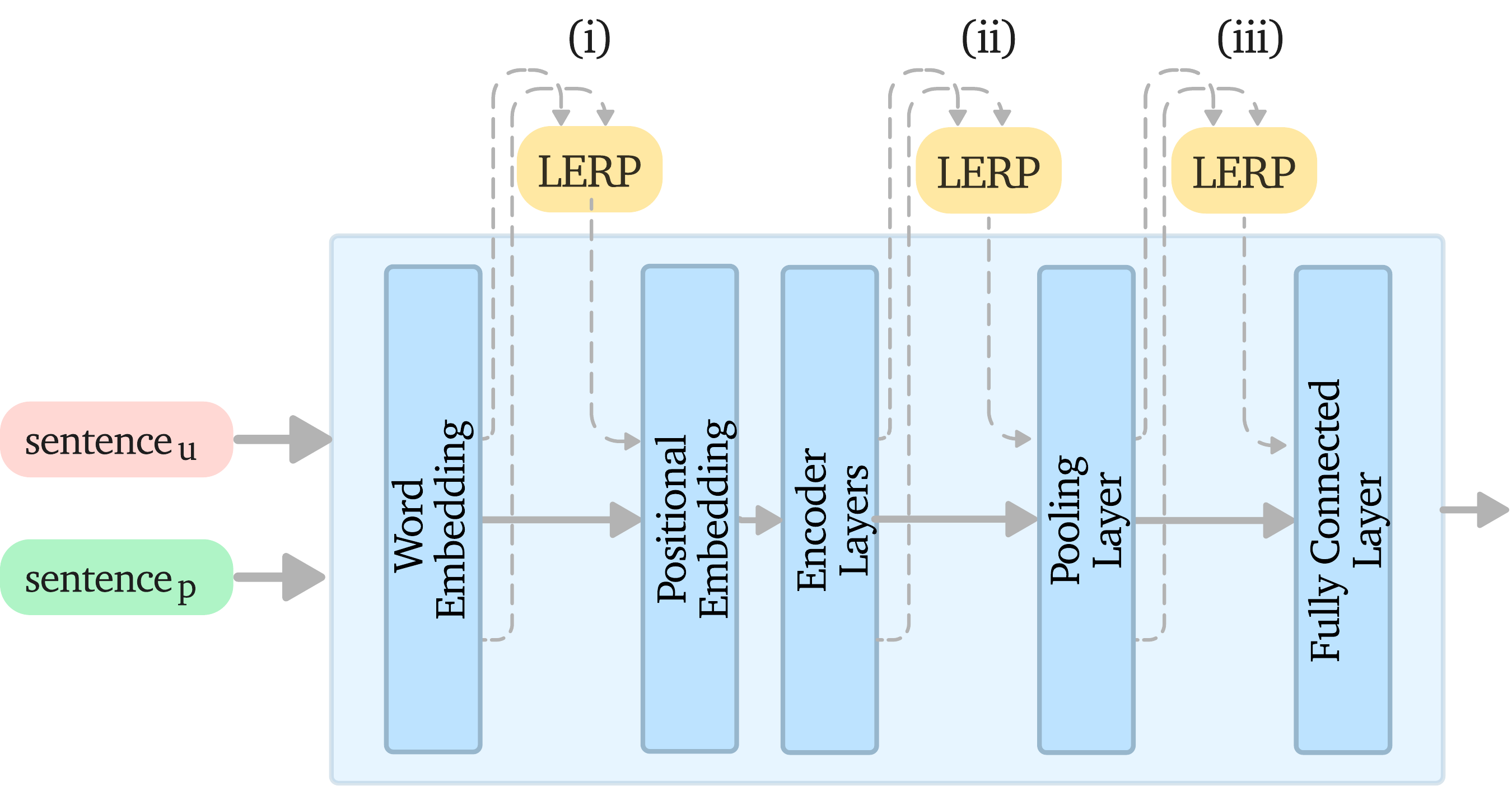}
        \caption{Examples of possible MixUp strategies by linear interpolation in LLM hidden representations: (i) word embedding, (ii) encoding, (iii) sentence embedding.}
        \label{fig:MixUp}
    \end{minipage}
    \hfill
    \begin{minipage}{0.52\textwidth}
        \tabcolsep=0.05cm
        \captionof{table}{Evaluation datasets source problem domains \& properties.}
        \label{tab:datasets}
        \begin{tabular}{c|c|c|c}
        \hline
        Dataset & Source/Target & \#Samples & \#Labels \\ \hline
        PubMed  &   Male/Female &   5757    &    14    \\
        arXiv   &     Old/New   &   7000    &    14    \\
        Movies  &     Wiki/IMDb &   1000    &    20
        \end{tabular}
    \end{minipage}
\end{figure}

\textbf{Evaluation Setup.}
We use three datasets from different textual categories: healthcare, academia, and entertainment. \textit{PubMed} is a medical dataset consisting of research articles from the PubMed\footnote{\url{https://pubmed.ncbi.nlm.nih.gov}} repository. The articles' subheadings denote the source and target domains, namely \textit{female} and \textit{male} patients. The labels represent different biological categories. \textit{ArXiv} provides a collection of research paper abstracts, where labels represent subjects into which to categorize the abstract. Here, the target and source domains are old and new articles scrapped from the ArXiv repository\footnote{\url{https://arxiv.org}}. The \textit{Movies} dataset contains a collection of movie summaries that may belong to different genres. The source and target domains for the movie overviews are Wikipedia and IMDb, respectively. \Cref{tab:datasets} shows an overview of the datasets' properties. Each dataset from both domains is split inot a 80\% training and 20\% testing using random partitioning. To simulate missing labels we discard 50\%, 70\%, and 90\% of the labels for each sample.

\textbf{Training Details.}
We use pretrained BERT \textit{bert-base-uncased}\footnote{\url{https://huggingface.co/bert-base-uncased}}. Specifically, we employ the \textit{BertForSequenceClassification} model and fine-tune it using the \textit{Trainer} class for easier reproducibility, efficient GPU memory utilization, and simplified workflow. All runs use batch size equal to 64, learning rate equal to 5e-5, and 12 training epochs. We set the $\lambda$, $\alpha$, $\beta$ to 1, 0.3, 0.3. We run each experiment thrice and report the average mAP score~\cite{8594067} on the target domain. Furthermore, we do not freeze BERT’s layers during the fine-tuning.

\textbf{Baselines.}
We compare \algo with supervised fine-tuning, no fine-tuning, unsupervised domain adaptation method AAD~\cite{ryu2020knowledge}, and an asymptotic upper bound using fully supervised fine-tuning without removing any labels. For AAD, we follow the authors' work, freezing BERT's layers and training for three epochs.

\begin{figure}[tb]
     \centering
     \begin{subfigure}[b]{0.3275\textwidth}
         \centering
         \includegraphics[width=\textwidth]{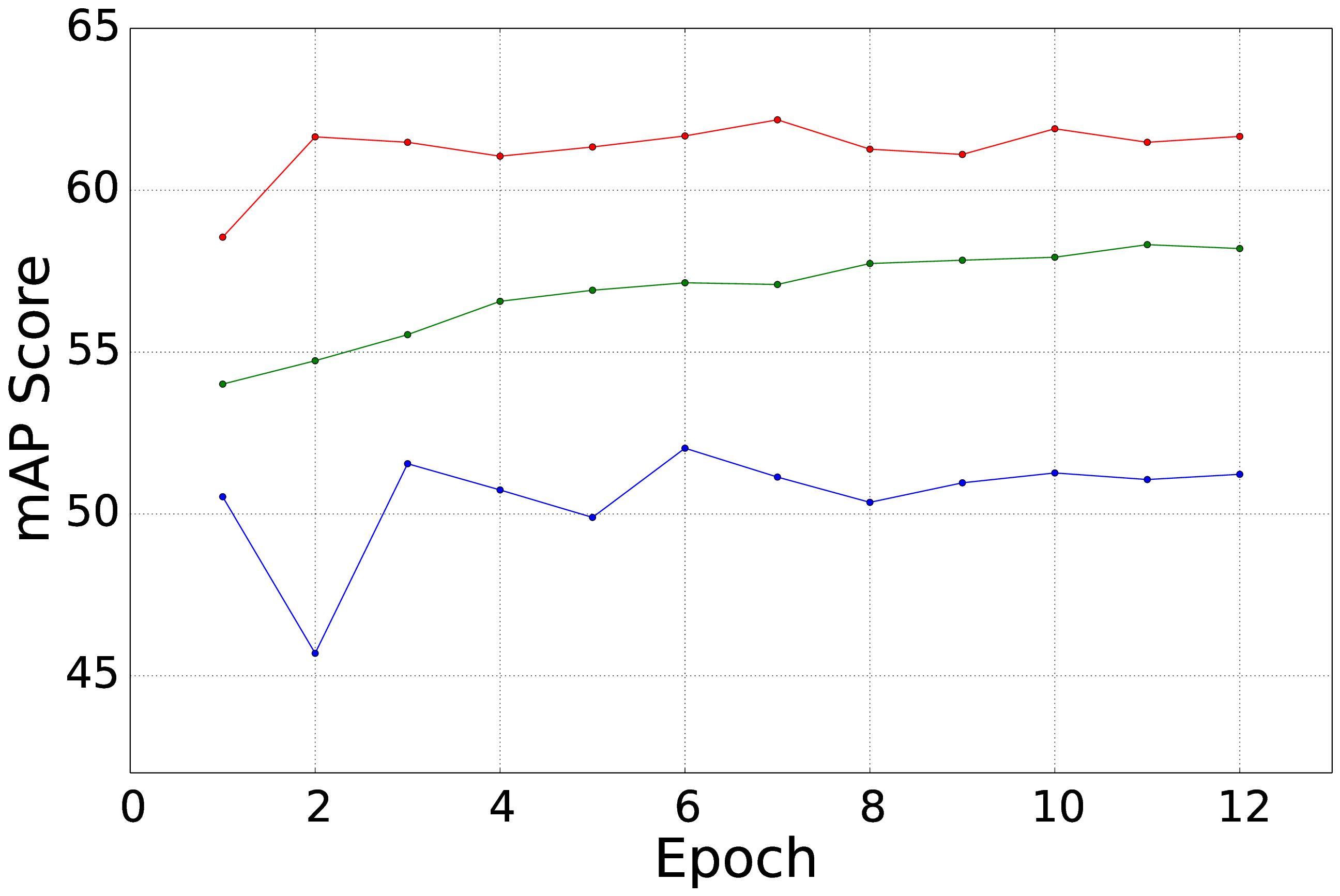}
         \caption{PubMed}
         \label{fig:map-a}
     \end{subfigure}
     \hfill
     \begin{subfigure}[b]{0.3275\textwidth}
         \centering
         \includegraphics[width=\textwidth]{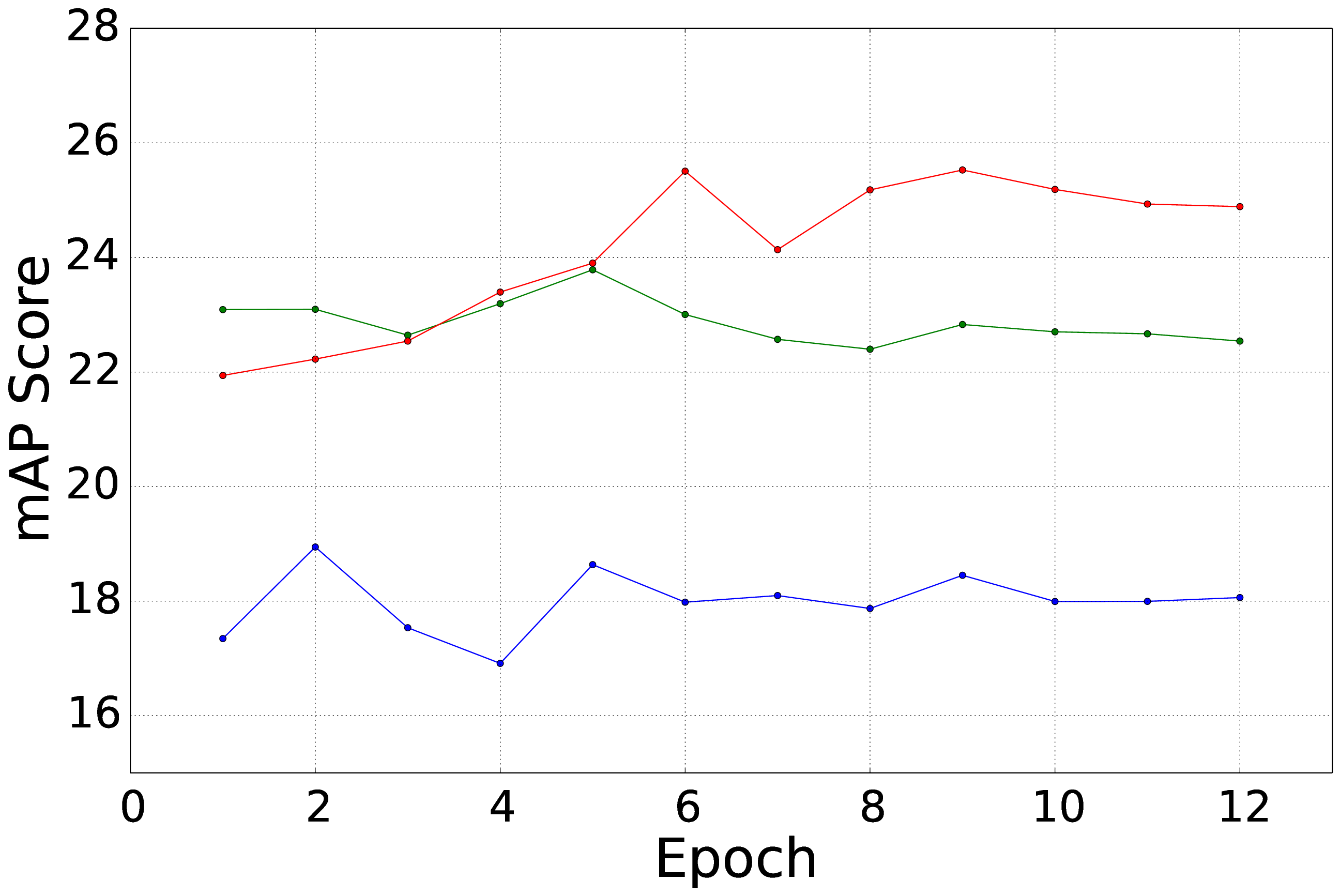}
         \caption{ArXiv}
         \label{fig:map-b}
     \end{subfigure}
     \hfill
     \begin{subfigure}[b]{0.3275\textwidth}
         \centering
         \includegraphics[width=\textwidth]{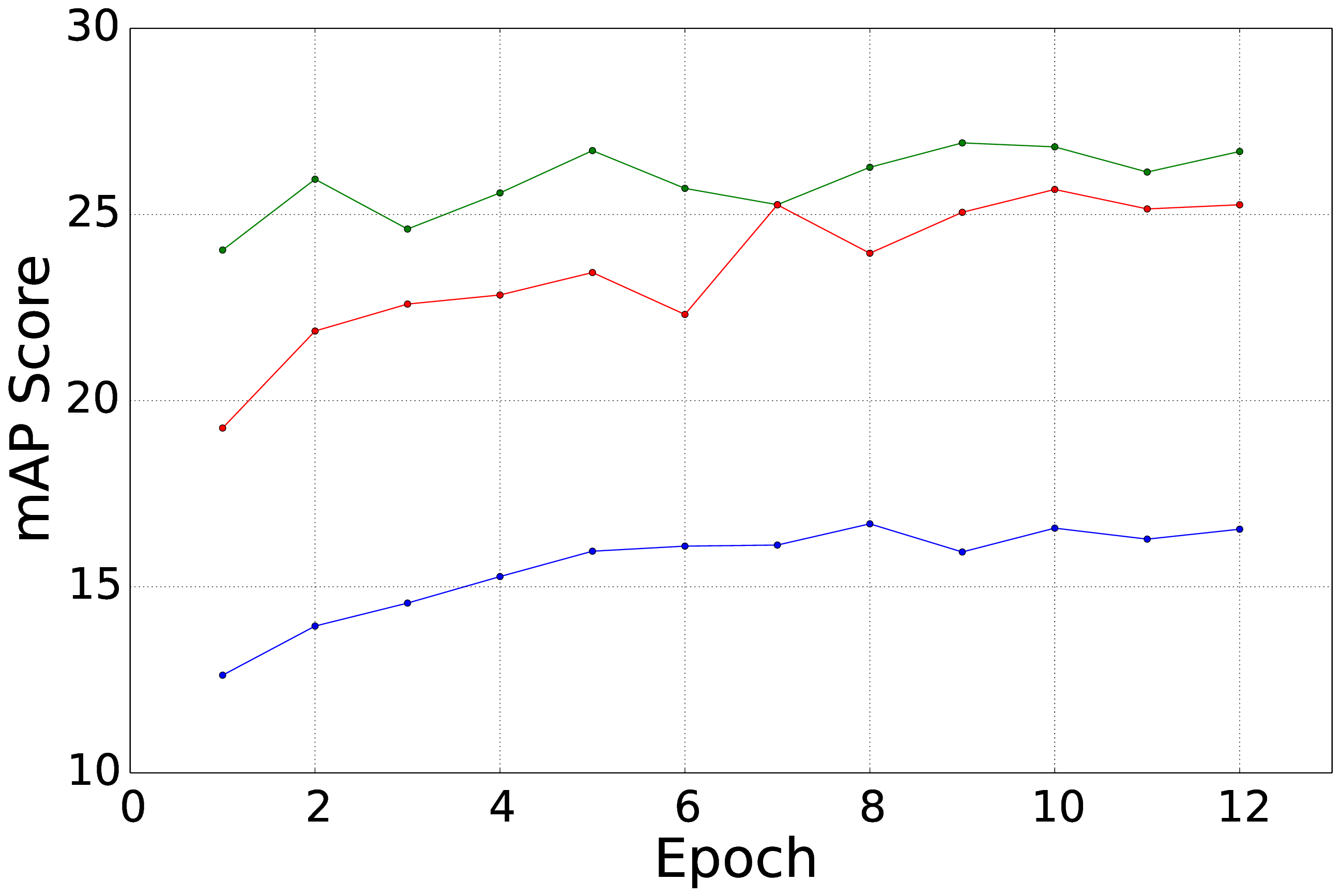}
         \caption{Movies}
         \label{fig:map-c}
     \end{subfigure}
        \caption{mAP scores\slash{}epoch for: supervised fine-tuning w\slash{} 50\% labels (blue), \algo w\slash{} 50\% labels (green), and supervised fine-tuning w\slash{} 100\% labels (red).}
        \label{fig:map}
\end{figure}

\textbf{Empirical Results.}
\autoref{tab:res-main} reports the mAP results of \algo compared to the baseline methods for different ratios of discarded labels per column. We observe that \algo consistently outperforms the supervised fine-tuning technique among all the dataset and label ratio combinations. Our method outperforms the unsupervised AAD in almost all cases, except for scenarios with only 10\% label availability, where it remains competitive. Consistent with expectations, the supervised fine-tuning, having all labels, outperforms our proposed algorithm. However, in the case of the Movies dataset, our method achieves marginally higher performance than fully supervised fine-tuning. The variation can be explained by the dataset's low sample count and the wide nature of the dataset. We argue that our approach's effectiveness is due to its incorporation of multiple interpolations and the generation of artificial samples, which contribute to higher results in such data settings. \autoref{fig:map} presents an overview of how the mAP score varies over the 12 epochs in the case of 50\% label removal for each dataset within a fixed seed. Our method's performance fits between that of the fully supervised fine-tuning with and without label removal. Here, in \autoref{fig:map-c}, we can better observe the phenomenon from ~\autoref{tab:res-main} in Movies. Our method achieved slightly better performance than supervised fine-tuning without label removal.

\begin{table}[tb]
\tabcolsep=0.065cm
\renewcommand{\arraystretch}{1.2}
\centering
\caption{mAP scores under different ratios of available labels per column for: target domain data of source model, regular fine-tuning (supervised), AAD\cite{ryu2020knowledge} (unsupervised), and \algo (semi-supervised).}
\label{tab:res-main}

\newcolumntype{g}{>{\columncolor[gray]{0.85}}c}

\begin{tabular}{l|ccgccccgc}
\hline
\multirow{3}{*}{Dataset} & \multicolumn{9}{c}{\begin{tabular}[c]{@{}c@{}}Method \&\\ Ratio Available Labels\end{tabular}} \\
                         \cline{2-10}
                         & \multicolumn{1}{c|}{None} & \multicolumn{4}{c|}{Fine-Tuning} & \multicolumn{1}{c|}{AAD} & \multicolumn{3}{c}{\algo} \\
                         & \multicolumn{1}{c|}{-} & 100\% & 50\% & 30\%  & \multicolumn{1}{g|}{10\%}   & \multicolumn{1}{c|}{-}      & 50\%   & 30\%   & 10\%   \\
                         \hline
PubMed                   & \multicolumn{1}{c|}{57.1} & 62.0\tiny{$\pm$\textbf{.}28} & 50.6\tiny{$\pm$0\textbf{.}40} & 45.3\tiny{$\pm$\textbf{.}36} & \multicolumn{1}{g|}{42.5\tiny{$\pm$\textbf{.}05}} & \multicolumn{1}{c|}{53.7\tiny{$\pm$\textbf{.}40}} & 58.9\tiny{$\pm$\textbf{.}48} & 58.2\tiny{$\pm$0\textbf{.}38} & 52.2\tiny{$\pm$\textbf{.}57} \\
arXiv                    & \multicolumn{1}{c|}{21.8} & 25.0\tiny{$\pm$\textbf{.}24} & 18.5\tiny{$\pm$\textbf{.}12} & 15.7\tiny{$\pm$\textbf{.}45} & \multicolumn{1}{g|}{13.7\tiny{$\pm$\textbf{.}19}} & \multicolumn{1}{c|}{21.5\tiny{$\pm$\textbf{.}03}} & 24.5\tiny{$\pm$\textbf{.}32} & 23.2\tiny{$\pm$\textbf{.}60} & 20.5\tiny{$\pm$\textbf{.}48} \\
Movies                   & \multicolumn{1}{c|}{20.7} & 24.7\tiny{$\pm$\textbf{.}39} & 15.8\tiny{$\pm$\textbf{.}55} & 13.2\tiny{$\pm$\textbf{.}09} & \multicolumn{1}{g|}{12.6\tiny{$\pm$\textbf{.}00}} & \multicolumn{1}{c|}{17.6\tiny{$\pm$\textbf{.}12}} & 26.5\tiny{$\pm$\textbf{.}21} & 26.2\tiny{$\pm$\textbf{.}48} & 26.0\tiny{$\pm$\textbf{.}36}
\end{tabular}
\end{table}

\textbf{Ablation Studies.}
To establish the effectiveness of \algo, we conduct three ablation studies. First, we analyze the impact of varying interpolation methods. Second, we compare our norm-based approach to the original log-based VPU. Third, we evaluate the efficiency of our sub-sampling technique.

\autoref{tab:res-ablation}{(i)} illustrates the differences in mAP scores resulting from the different interpolation methods, as opposed to the base word embedding interpolation. The reference mAP scores for the comparative analysis are those from \autoref{tab:res-main}. In particular, the encoding and sentence embeddings both show analogous scores, having slightly lower performance compared to the word embeddings in the PubMed and ArXiv datasets. In contrast, using encoding or sentence-based interpolations in the Movies dataset results in a marginal mAP score increase. This finding highlights the adaptability of the interpolation mechanism, indicating its potential for customization to the specifics of individual datasets. The Movies dataset constitutes a unique case study, where our method is validated under conditions characterized by a limited number of samples in a wide dataset.

\autoref{tab:res-ablation}{(ii)} shows the changes in mAP scores obtained by our method using a log-based variational approach. Exclusively, these scores display a decline in performance across all three datasets and even fall below those achieved by supervised fine-tuning with label removal. We argue that the complexities imposed by the internal representations of textual data make log-based methods overly strict with respect to deviations from learning, leading to a pronounced amplification of errors and, consequently, to poorer performance. In contrast, our proposed method consistently outperforms the baselines, with competitive results in the few cases where it does. This contrast highlights the robustness and validity of our approach over logarithmic variational strategies.

\begin{table}[tb]
\tabcolsep=0.1cm
\renewcommand{\arraystretch}{1.2}
\centering
\caption{Relative changes in mAP score for \algo's \autoref{tab:res-main} results when changing (i) the word-level MixUp to encoding\slash{}sentence-level and (ii) the Norm-based loss to the Log-based formulation (negative percentage means score decrease).}
\begin{tabular}{l|cccccc}
\hline
\multirow{3}{*}{Dataset} & \multicolumn{6}{c}{\begin{tabular}[c]{@{}c@{}}Ablation \&\\ Ratio Available Labels\end{tabular}}      \\ \cline{2-7}
                         & \multicolumn{3}{c|}{(i) Encoding/Sentence MixUp}         & \multicolumn{3}{c}{(ii) Log-based Loss}   \\
                         & 50\%       & 30\%       & \multicolumn{1}{c|}{10\%}       & 50\%       & 30\%       & 10\%            \\ \hline
PubMed                   & -0.018\%   & -0.167\%   & \multicolumn{1}{c|}{-0.937\%}   & -20.94\%   & -21.80\%   & -16.73\%        \\
arXiv                    & -0.497\%   & -0.455\%   & \multicolumn{1}{c|}{-1.234\%}   & -24.94\%   & -23.34\%   & -18.16\%        \\
Movies                   & +0.547\%   & -0.966\%   & \multicolumn{1}{c|}{-0.935\%}   & -39.76\%   & -44.47\%   & -46.75\%
\end{tabular}
\label{tab:res-ablation}
\end{table}

\autoref{tab:res-data} illustrates the differences in training time and mAP score associated with different batch sampling strategies. In the first version of the \algo, we use a standard unweighted sampler, where the batches have a higher risk of being imbalanced, leading to a poorer performance, even if the training time is low. In the second version, we use nested batching to overcome the imbalance, pairing a large batch of unlabeled samples with smaller batches associated with positive samples for each specific label. Despite showing comparable results to our suggested method, nested batching significantly increases training time. In our final approach, we use a cycle sampler whereby each batch contains at least one positive sample from every label. This adapted sampling technique maintains quality predictions while substantially reducing the training duration.

\begin{table}[tb]
\tabcolsep=0.065cm
\renewcommand{\arraystretch}{1.2}
\centering
\caption{Training time and mAP score w\slash{} 50\% labels for: unweighted sampler, and nested batching. \algo achieves training time similar to the unweighted sampler with performance on the level of nested batching.}
\label{tab:res-data}
\begin{tabular}{l|cc|cc}
\hline
\multicolumn{1}{c|}{\multirow{2}{*}{Dataset}} & \multicolumn{2}{c|}{Training time (s)} & \multicolumn{2}{c}{mAP Score}     \\
\multicolumn{1}{c|}{}                         & Unweighted Sample  & Nested Batch  &  Unweighted Sample & Nested Batch \\ \hline
PubMed       & 1793.45 & 2607.62 &58.13 & 58.86\\
arXiv        &  1553.97    &  2796.19   &   22.92 &  23.61   \\
Movie Genres &  369.54    &  631.78    &  25.36 &   25.22
\end{tabular}
\end{table}

\section{Conclusion}

To tackle the domain shift in classifying unlabeled text data, we propose a semi-supervised domain adaptation approach, \algo, specifically for LLM-based multi-label classifiers. By incorporating a novel variational loss, MixUp regularizer, and label-balanced sampling, \algo effectively exploits the limited positively labeled and abundant unlabeled text data during domain adaptation. We evaluate \algo against partial-supervised and unsupervised methods across three datasets, accounting for various label availability scenarios, demonstrating its efficiency. Importantly, our method achieves higher mAP scores, exceeding unsupervised and partial-supervised approaches by 19.9\% and 52.2\%, respectively. Our results showcase \algo's effectiveness and resilience of in enhancing multi-label classification tasks in the presence of domain shifts and scarce labels.

\textbf{Acknowledgements}
This work has been supported by the Spoke ``FutureHPC \& BigData'' of the ICSC - Centro Nazionale di Ricerca in ``High Performance Computing, Big Data and Quantum Computing'', funded by EU - NextGenerationEU and the EuPilot project funded by EuroHPC JU under G.A. 101034126.

\bibliographystyle{splncs04}
\bibliography{bibliography}

\end{document}